# Orthogonally Initiated Particle Swarm Optimization with Advanced Mutation for Real-Parameter Optimization

Indu Bala
School of Computer and Mathematical Sciences
The University of Adelaide South, Australia, Australia
indu.bala@adelaide.edu.au

Dikshit Chauhan
Department of Mathematics and Computing, Dr. B.R. Ambedkar National Institute of Technology, Jalandhar, Punjab, India, 144008
dikshitchauhan608@gmail.com

Lewis Mitchell
School of Computer and Mathematical Sciences
The University of Adelaide South, Australia, Australia
lewis.mitchell@adelaide.edu.au

## ABSTRACT
This article introduces an enhanced particle swarm optimizer (PSO), termed **O**rthogonal **PSO** with **M**utation (OPSO-m). Initially, it proposes an orthogonal array-based learning approach to cultivate an improved initial swarm for PSO, significantly boosting the adaptability of swarm-based optimization algorithms. The article further presents archive-based learning strategies, dividing the population into regular and elite sub-groups. Each sub-group employs distinct learning mechanisms. The regular group utilizes efficient learning schemes derived from three unique archives, which categorize individuals based on their quality levels. Additionally, a mutation strategy is implemented to update the positions of elite individuals. Comparative studies are conducted to assess the effectiveness of these learning strategies in OPSO-m, evaluating its optimization capacity through exploration-exploitation dynamics and population diversity analysis. The proposed OPSO-m model is tested on real-parameter challenges from the CEC 2017 suite in various dimensional search spaces. OPSO-m exhibits distinguished performance in the precision of solutions, rapidity of convergence, efficiency in search, and robust stability, thus highlighting its superior aptitude for resolving optimization problems.

## CCS CONCEPTS
• Computing methodologies → Particle Swarm Optimization

## KEYWORDS
Particle Swarm Optimization, Swarm Intelligence, Metaheuristic Optimization, Orthogonal Initialization

## 1 Introduction

Particle Swarm Optimization (PSO) is a widely utilized swarm intelligence approach [1] for addressing optimization challenges. Inspired by the swarm behavior of social animals such as bird flocking, PSO operates via a swarm of particles navigating an n-dimensional search space, with each particle's position and velocity representing a potential solution to the optimization problem. The global optimum is sought through iterative updates to each particle's velocity and position, governed by the following equations:

$$v_i^{t+1} = \omega v_i^t + c_1 r_1 (Pbest_i^t - x_i^t) + c_2 r_2 (Gbest^t - x_i^t) \quad (1)$$
$$x_i^{t+1} = x_i^t + v_i^{t+1} \quad (2)$$

Where $t$ denotes the iteration number. $v_i^t$ and $x_i^t$ represent the velocity and position of the ith particle, respectively. The inertia weight is denoted by $\omega$, while $c_1$ and $c_2$ are acceleration coefficients [2]. The algorithm also involves two random vectors, $r_1$ and $r_2$, generated within the range [0, 1]. The terms $Pbest_i^t$ and $Gbest^t$ refer to the personal best solution of $ith$ particle and the global best solution found by the entire swarm, respectively. $c_1 r_1 (Pbest_i^t - x_i^t)$ and $c_2 r_2 (Gbest^t - x_i^t)$ present the cognitive and social components of the algorithm [3].

PSO has been updated to solve various optimizations tasks [4-7], Nonetheless, PSO may encounter difficulties with optimization problems featuring numerous local optima or high dimensions, which can lead to premature convergence [4]. To address this, the study of orthogonal learning, advanced learning schemes, and mutation techniques into the PSO algorithm results in a more balanced approach, ensuring effective convergence and thorough exploration of the search space.

Our significant contributions in this article include:
1. Integrating an orthogonal array with PSO for effective population initialization.
2. Introducing three learning schemes, each anchored in distinct archival sources, beneficial for regular individuals, and applying mutation strategies to enhance the positioning of elite individuals. This promotes a balance between exploitation and exploration and helps to avoid local optima.

## 2 Method

In this section, we will explore the learning methodology implemented in the OPSO-m approach and how it contributes to improved performance, with detailed discussions provided in the following sections.

### 2.1 Orthogonal-Based Initialization Strategy

This section details the implementation of an orthogonal array (OA) in establishing the initial swarm for the proposed OPSO-m model. This approach facilitates a comprehensive and even scrutiny of the search domain, assuring a balanced distribution across the OPSO-m individuals [8].

An OA is conceptualized as a matrix filled with elements from a set $S$ that contains $\alpha$ levels (for instance, $\{1, 2, ..., \alpha\}$). This matrix is structured so that for any integer $z$, selecting any $z$ columns from the array will show every possible z-tuple of symbols, with each tuple occurring equally throughout the rows when considering only those columns. Mathematically, an orthogonal array can be denoted as $A_\beta(\alpha^\delta) = [a_{j,k}]_{\beta \times \delta}$ where $a_{j,k}$ represents the symbol assigned to the $kth$ factor in the $jth$ combination, and each $a_{j,k}$ belongs to the set $\{1, 2, ..., \alpha\}$. Here, $A$ signifies the arrangement akin to a Latin square, $\beta$ indicates the number of combinations, and $\delta$ denotes the number of factors.

Upon initializing the orthogonal array, we adapt the continuous search space of PSO to a discrete format because the OA's design is intrinsically meant for discrete spaces [9]. We achieve this using the following transformation:

$$x_{j,k} = a_{j,k} \times \left(\frac{x_{ub} - x_{lb}}{\max(A) - \min(A)}\right) + x_{lb}, j = 1, 2, ..., \beta, k = 1, 2, ..., \delta. \quad (3)$$

This formula is instrumental in mapping the discrete entries from the OA to continuous points within the prescribed lower ($x_{lb}$) and upper ($x_{ub}$) limits of the search space [10]. By executing this transposition process on each element of the OA, we determine the initial continuous positions $x_{j,k}$ for the OPSO-m swarm, setting a strong foundation for the exploration process.

### 2.2 Mechanism of proposed archives

In this section, following the orthogonal initialization of the algorithm, we sort the entire swarm in ascending order of fitness values and divide into two distinct subgroups: regular and elite. Each of these subgroups employs different learning mechanisms.

The elite subgroup undergoes updates through a mutation strategy, enhancing exploration capabilities in the mid-stages of the optimization process. Conversely, the regular subgroup is updated through a three-archive system, which accumulates and utilizes the collective information of promising individuals from the entire swarm. The integration of these archives through three learning schemes facilitates information sharing within the swarm, thereby improving the exploitation ability of the algorithm, described as follows:

1. **First archive $\Phi^t$**: This archive compiles the personal best performances of the top-performing individuals, expressed as $\Phi^t = \{\Phi_1^t, \Phi_2^t, ..., \Phi_{n/2}^t\}$ where n/2 is length of this archive.

2. **Second archive $\psi^t$**: This archive is dedicated to storing promising personal bests from the entire swarm or population, with its maximum capacity equal to the total population count. It is represented as $\psi^t = \{\psi_1^t, \psi_2^t, ..., \psi_n^t\}$.

3. **Third archive $\chi^t$**: This archive is utilized for collecting the global bests that show promise, the length of this archive is on par with the swarm size, symbolized by $\chi^t = \{\chi_1^t, \chi_2^t, ..., \chi_n^t\}$.

These archival methods are particularly tailored to refine the position of the regular individuals, making up the second segment of the population. To support this process, three distinct learning schemes are deployed, specifically designed for addressing minimization challenges. For the sake of this explanation, let us consider $\Phi_p^t, \psi_q^t, \chi_s^t$ to be representatives selected from each of the respective archives, where $p$, $q$, and $s$ are indices within the ranges $\{1, 2, ..., \frac{n}{2}\}, \{1, 2, ..., n\}, \{1, 2, ..., n\}$ chosen at random.

**Scheme 1:** When the fitness of the selected individual from the first archive $\Phi_p^t$, is superior to that of individuals from the second and third archives $\psi_q^t$, and $\chi_s^t$ respectively, the regular individual $j$ updates its velocity as follows:

$$v_{r,j,k}^{t+1} = r_{1,k}^t \times v_{r,j,k}^t + r_{2,k}^t \times \left(\Phi_{p,k}^t - x_{r,j,k}^t\right) + r_{3,k}^t \times \left(Gbest^t - x_{r,j,k}^t\right), j = \{1, 2, ..., n/2\}, k = \{1, 2, ..., d\} \quad (4)$$

where $r_{1,k}^t$ are random numbers between 0 and 1. The term $\Phi_{p,k}^t$ denotes the location of the $kth$ dimension in $\Phi_p^t$, $v_{r,j,k}^t$ and $x_{r,j,k}^t$ are velocity and position of $kth$ variable of $jth$ regular individual.

**Scheme 2**: If the fitness value of an individual from the second archive $\psi_q^t$ is better than the first $\Phi_p^t$ and third archives $\chi_s^t$ then the regular individual $j$ updates its velocity as follows:

$$v_{r,j,k}^{t+1} = r_{1,k}^t \times v_{r,j,k}^t + r_{2,k}^t \times \left(\psi_{q,k}^t - x_{r,j,k}^t\right) + r_{3,k}^t \times \left(Gbest^t - x_{r,j,k}^t\right), j = \{1, 2, ..., n/2\}, k = \{1, 2, ..., d\} \quad (5)$$

where term $\psi_{q,k}^t$ denotes the location of the $kth$ dimension in $\psi_q^t$ like previous scheme.

**Scheme 3:** When the fitness of the individual from the third archive $\chi_s^t$ surpasses that of individuals from the first and second archives, the regular individual $j$ updates its velocity as:

$$v_{r,j,k}^{t+1} = r_{1,k}^t \times v_{r,j,k}^t + r_{2,k}^t \times \left(\chi_{s,k}^t - x^t_{r,j,k}\right) + r_{3,k}^t \times \left(Gbest^t - x^t_{r,j,k}\right), \; j = \{1,2,\dots,n/2\}, k = \{1,2,\dots,d\} \quad (6)$$

where term $\chi_{s,k}^t$ denotes the location of the $kth$ dimension in $\chi_s^t$. These schemes explore how individuals within a population-based optimization algorithm can learn from different archival information to update their search for the optimum solution [11].

### 2.3 Mutation Strategy

The mutation strategy plays a pivotal role in refining the positions of the elite subgroup within the OPSO-m model. The elite subgroup, representing one half of the overall population post-division, benefits from this strategy which facilitates a dynamic learning environment. Specifically, individuals within the elite subgroup are not only influenced by their own historical best positions but are also guided by the positions of two peers randomly chosen from their cohort. This approach is known in literature as the "$best - rand$ to current" mutation strategy [6]. The foundational mathematical expression for this mutation is described as follows:

$$x_{e,j,k}^{t+1} = x_{e,j,k}^t + \Delta_{1,k}^t \times \left(\Phi_{j,k}^t - x_{e,j,k}^t\right) + \Delta_{2,k}^t \times \left(x_{g,j}^t - x_{h,j}^t\right), \; j = \{1,2,\dots,n/2\}, k = \{1,2,\dots,d\} \quad (7)$$

Where $\Delta_{1,k}^t$ and $\Delta_{2,k}^t$ are scaling factor between 0 and 1, and $g$ and $h$ represent two distinct randomly selected elements from the elite subgroup.

## 3 Experimental Findings and Analysis

To evaluate the efficiency of the proposed OPSO-m model, we applied it to the benchmark problems from the CEC2017 test suite [12]. These problems are sorted into four categories: unimodal ($Fno_1 - Fno_3$), multimodal ($Fno_4 - Fno_{10}$), hybrid ($Fno_{11} - Fno_{20}$), and composite ($Fno_{21} - Fno_{30}$), each with a search range of $[-100, 100]^d$. The performance of OPSO-m was rigorously tested 25 times per problem on different dimensions to determine its global search capabilities. These evaluations were benchmarked against six other algorithms: Particle Swarm Optimization (PSO) [1], Ant Colony Optimization (ACO) [13], Gravitational Search Algorithm (GSA) [14], Differential Evolution (DE) [15], Covariance Matrix Adaptation Evolution Strategy (CMA-ES) [16], and Reinforcement Learning-based Memetic Particle Swarm Optimizer (RLMPSO) [17], with parameter settings consistent with those reported in the cited literature. The computational tests were performed on a Windows 10 system utilizing MATLAB (2019). The results were compiled based on the *'best-of-run'* error values for each algorithm, where the error is the absolute discrepancy between the best-obtained value and the known optimal value.

### 3.1 Results Discussion on Implemented Strategies

The effectiveness of the implemented strategies in the OPSO-m model—orthogonal initialization, archive utilization, and mutation is evaluated using convergence graphs depicted in Figure 1.

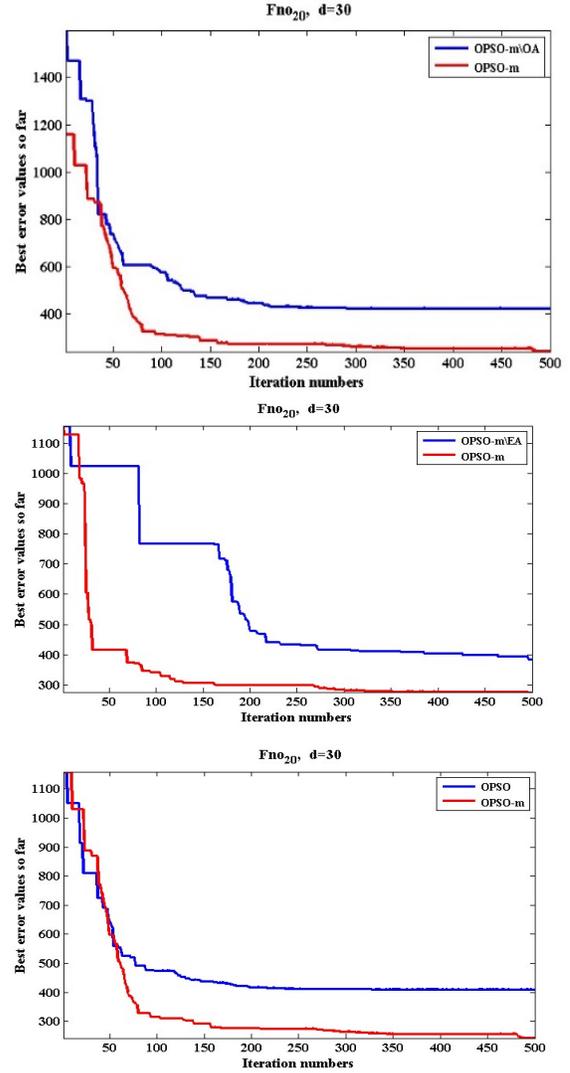

**Figure 1: Comparative Effectiveness of Orthogonal Array-Based Initialization, Archive Utilization and Mutation Strategies respectively.**

Figure 1 focuses on hybrid function $Fno_{20}$ at 30-dimensional level selected for their complexity and solution intricacies. Here, OPSO-m\OA represents our model without the OA integrated, OPSO-m\EA (without archives), and OPSO (without mutation). Figure 1 clearly demonstrates that OPSO-m with these implemented strategies is evidenced by its more stable convergence trajectory and reduced error margins. These convergence plots distinctly highlight the OPSO-m model's increased efficiency, a direct result of these strategic integrations. Notably, the model achieves a faster rate of initial convergence, allowing for quick identification of promising solution regions. Additionally, the uniform achievement of lower error values throughout successive iterations indicates a refined and efficient optimization process, leading to an effective and consistent search for the optimal solutions.

Figure 2 provides a detailed convergence analysis of the multimodal function $Fno_{20}$ across 30 and 50 dimensions. Each plots the best error values so far against the number of iterations for a variety of optimization algorithms, including GSA, ACO, DE, CMA-ES, PSO, RLMPSO, and OPSO-m. From the figure, it is apparent that the OPSO-m algorithm demonstrates superior convergence behavior across both dimensions when compared to the other algorithms. This is particularly noticeable in the way OPSO-m's curve descends swiftly and levels out at a lower best error value early in the iteration process, which suggests that it is able to find promising areas of the search space more quickly and efficiently. Furthermore, OPSO-m maintains this advantage throughout the iterative process, suggesting robustness and a strong ability to fine-tune solutions to approach or reach the global optimum.

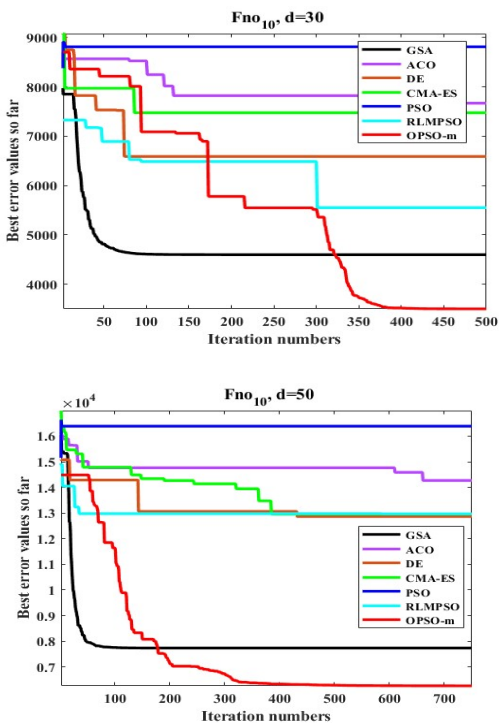

**Figure 2: Analysis of Convergence for Multimodal Function No. 10 Across Dimensions 30, and 50 respectively.**

For thorough validation of the OPSO-m model's robustness, both dimensional sensitivity analysis and diversity trends were explored. However, due to space limitations, the full convergence results for all CEC2017 test suite functions, along with sensitivity and diversity analyses, are stored in the supplementary material on the GitHub repository: https://github.com/InduBala-Y/OPSO-m-.

## 4 Conclusion

In summary, OPSO-m represents a significant advancement in particle swarm optimization, incorporating orthogonal learning for initial population generation and employing a dual-subpopulation strategy. This strategy differentiates learning mechanisms for the regular subgroup, based on elite archives, and applies mutation strategies for the elite subgroup. Such an approach ensures a balanced exploration and exploitation, while achieving robust convergence. The effectiveness of each component in OPSO-m has been rigorously evaluated, showing a substantial improvement in the algorithm's performance. When benchmarked against six leading algorithms, OPSO-m demonstrated superior capabilities in solution accuracy, convergence rates, and computational efficiency. The findings of this study confirm OPSO-m's adaptability and its potential for wide-ranging applications in future optimization challenges.